\documentclass[letterpaper, 10 pt, conference]{ieeeconf}  

\IEEEoverridecommandlockouts                              

\usepackage{graphics} 
\usepackage{epsfig} 
\usepackage{mathptmx} 
\usepackage{times} 
\usepackage{amsmath} 
\usepackage{amssymb} 

\usepackage[caption=false, font=footnotesize]{subfig}
\usepackage{url}

\newcommand{\FigRef}[1]{Fig.~\ref{#1}}
\newcommand{\TabRef}[1]{Tab.~\ref{#1}}
\renewcommand{\vec}{\mathbf}
\newcommand{\mat}{\mathbf}

\title{\LARGE \bf
Analytic Estimation of Region of Attraction of an LQR Controller for Torque Limited Simple Pendulum
}

\author{Lukas Gross$^{1,2}$,
Lasse Maywald$^{1}$,
Shivesh Kumar$^{1}$,
Frank Kirchner$^{1}$, and
Christoph L\"uth$^{2}$
\thanks{This work was supported by the VeryHuman (Grant Number: 01IW20004) project funded by the German Aerospace Center (DLR) with federal funds from the Federal Ministry of Education and Research (BMBF).}
\thanks{$^{1}$ Robotics Innovation Center, DFKI, 28359 Bremen, Germany.}%
\thanks{$^{2}$ Cyber-Physical Systems, DFKI, 28359 Bremen, Germany.}%
\thanks{ }%
\thanks{\textbf{\copyright~2022 IEEE.  Personal use of this material is permitted. Permission from IEEE must be obtained for all other uses, in any current or future media, including reprinting/republishing this material for advertising or promotional purposes, creating new collective works, for resale or redistribution to servers or lists, or reuse of any copyrighted component of this work in other works.}}%
}

\begin{document}

\maketitle
\thispagestyle{empty}
\pagestyle{empty}

\begin{abstract}
Linear-quadratic regulators (LQR) are a well known and widely used tool in control theory for both linear and nonlinear dynamics. For nonlinear problems, an LQR-based controller is usually only locally viable, thus, raising the problem of estimating the region of attraction (ROA). The need for good ROA estimations becomes especially pressing for underactuated systems, as a failure of controls might lead to unsafe and unrecoverable system states. Known approaches based on optimization or sampling, while working well, might be too slow in time critical applications and are hard to verify formally. In this work, we propose a novel approach to estimate the ROA based on the analytic solutions to linear ODEs for the torque limited simple pendulum. In simulation and physical experiments, we compared our approach to a Lyapunov-sampling baseline approach and found that our approach was faster to compute, while yielding ROA estimations of similar phase space area.
\end{abstract}

\section{Introduction}
\label{sec:int}
It is well known that systems with linear dynamics can be stabilized optimally w.r.t. a quadratic cost using a Linear-Quadratic Regulator (LQR) controller~\cite{book:AM90},~\cite{inbook:Tedrake22}. While they are straightforward to construct, LQR-controllers and the variant TVLQR proved to be useful tackling problems involving nonlinear dynamics. Possible applications range from stabilizing a fixed point using the linear approximations LQR to stabilizing full trajectories by computing multiple locally viable LQR-controllers along the trajectory~\cite{inproceedings:LT04, tedrake2010lqr}. 

A sub-problem arising from LQR-methods is the estimation of the respective Region of Attraction (ROA), i.e. the subset of the statespace in which the system is guaranteed to evolve towards the desired final state~\cite{book:Khalil02}. This ROA is not only constrained by the precision of the linear approximation that is used, but also by underactuation due to unactuated degrees of freedom and force or torque limitations.

Commonly, ROA estimation uses the Lyapunov criteria for stability. As known in the study of dynamical~\cite{article:HF96} as well as power systems~\cite{article:Gless66, article:KBP85, book:Pai12}, for an uncontrolled system the total energy is usually a very good candidate for a Lyapunov function, as system losses ensure its time-derivative to be negative. However, this does not translate to controlled systems, as a controller might induce additional energy into the system. Because of this, the need to construct suitable candidates for Lyapunov functions arises.

For an LQR-controller the naive ansatz is the cost-to-go, which is computed during the LQR-controllers derivation, as it is at least locally valid around the origin of the nonlinear dynamics. More refined methods include using sums-of-squares (SOS) or convex optimization~\cite{article:Johansen00, article:CGTV05, inproceedings:TD02}. Any sublevel set of this function, in which all Lyapunov conditions hold, when force/ torque limitations are taken into account, is an inner estimation of the actual ROA. The largest sublevel set can be determined via sampling~\cite{article:NB16, article:MMT21} or, again, by using SOS optimization~\cite{inproceedings:VTG09, article:TM10}.

The methods portrayed above, while working well in practice, have two main weaknesses. Firstly, cost-to-go is quadratic and thus the resulting sublevel sets are of ellipsoidal form. Lyapunov functions derived by SOS optimization can yield different shapes, but in practice, it is often quadratic as well. There is, however, no reason to expect the actual ROA to be ellipsoidal. Hence, the resulting ROA potentially severly underestimates the actual region. Secondly, ROA analysis is of special interest in safety-critical applications, where, more often than not, formal verification is essential, but since optimization is an intricate numerical approach and sampling depends on statistics, they tend to be hard to grasp using formal methods~\cite{article:MMT21}.

\begin{figure}[!t]
\centering
\hfill
\subfloat[\label{fig:sysa}]{\includegraphics[height=0.5\linewidth]{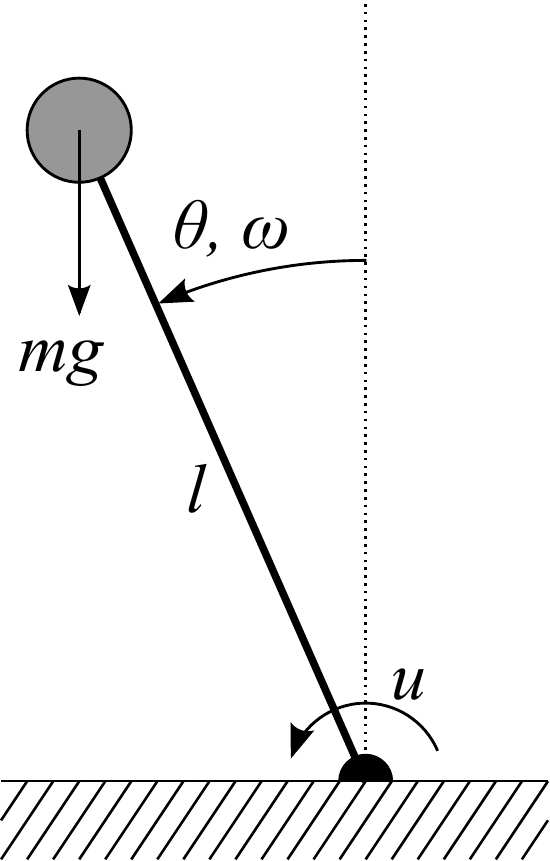}}
\hfill
\subfloat[\label{fig:sysb}]{\includegraphics[height=0.5\linewidth]{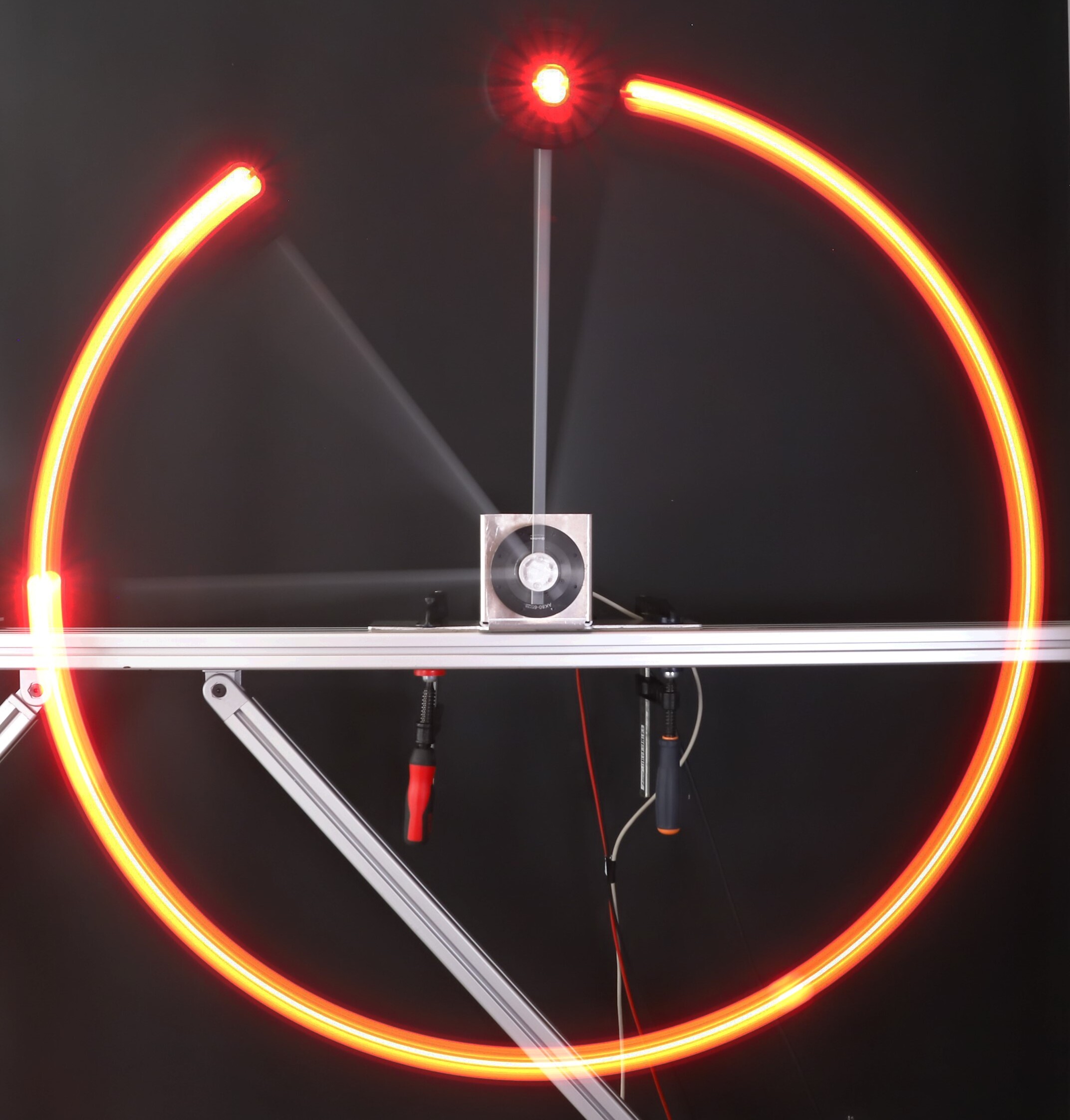}}
\hfill
\caption{(a) schematic of the pendulum, (b) long exposure shot of the physical system}
\label{fig:sys}
\end{figure}

\paragraph*{Contribution}
As a first step towards an alternative which does not suffer from the above shortcomings, we present an analytic approach to ROA estimations for the stabilization of a torque limited simple pendulum in its unstable fixed point (upright configuration). We chose a pendulum for its simple yet nonlinear dynamics, which exhibit features of underactuation. We found that our approach covered a ROA with area similar to that of the baseline, with a tendency to outperform it for tight torque limitations. In addition, we found that computational time required to compute the ROA with our proposed analytic method is multiple orders of magnitude faster than the baseline method. The approach has also been validated on a physical setup (see Fig.~\ref{fig:sysb}) in experiments.

\paragraph*{Organization}
In Section~\ref{sec:dyn}, the closed loop dynamics of an LQR-controlled torque limited pendulum are described in detail. In Section~\ref{sec:ana}, we use an analytic solution to the initial value problem of the linearized system to derive sufficient conditions for a system state to be part of the ROA. These findings are evaluated in Section~\ref{sec:exp} in simulation as well as on a physical setup and their performance is compared to a Lyapunov-sampling approach as baseline. Section~\ref{sec:con} concludes the paper and highlights the future work.
 
\section{Closed Loop Dynamics of an LQR-controlled Torque Limited Simple Pendulum}
\label{sec:dyn}
Consider a pendulum consisting of a rigid arm of length $l$ with an actuator at its axis and a weight of mass $m$ attached to its end. For simplicity and without loss of generality, consider the arm to be weightless. The equation of motion for such a pendulum and its first order approximation around the upright unstable fixed point $(\theta = \omega = 0)$ are given as
\begin{align}
\ddot{\theta} &= \frac{1}{ml^2} \left(mgl \sin(\theta) - b\dot\theta + u\right) \label{eq:dyn} \\
&\simeq \frac{1}{ml^2} \left(mgl \theta - b\dot\theta + u\right), \label{eq:lindyn}
\end{align}
where $g$ is the gravitational acceleration, $b$ a damping constant and $u$ the torque actuation. The direction of angle $\theta$, angular velocity $\omega = \dot{\theta}$ and the torque are defined as in \FigRef{fig:sys}(a). Defining the variables that way, the origin of the phase space corresponds to the upright fixed point. With $\vec{x}=(\theta, \dot{\theta})^T$ the linear approximation can be decomposed into a set of first order ordinary differential equations (ODE) as
\begin{equation}
\dot{\vec{x}} = \underbrace{\left(\begin{array}{cc} 
0 & 1 \\ 
\frac{g}{l} & -\frac{b}{ml^2} \\
\end{array}\right)}_{\mat{A}} \vec{x} + \underbrace{\left(\begin{array}{c} 
0  \\ 
\frac{1}{ml^2} \\
\end{array}\right)}_{\mat{B}} u.
\end{equation}

For a system of the form $\dot{\vec{x}} = \mat{A} \vec{x} + \mat{B}u$ and a quadratic cost function of the form $\int_0^\infty \left(\vec{x}^T\mat{Q}\vec{x}+uRu\right) dt$, the LQR-method yields an optimal controller $u(\vec{x}) = -\mat{K}\vec{x}$, which stabilizes the systems origin $\vec{x}^* = (0, 0)^T$, where $\mat{K} = \mat{R}^{-1}\mat{B}^T\mat{S}$ and $\mat{S}$ is the solution of the algebraic Riccati equation $0=\mat{SA}+\mat{A}^T\mat{S}-\mat{SBR}^{-1}\mat{B}^T\mat{S}+\mat{Q}.$ In this case, $\mat{K}$ is of the form $\mat{K} = \left(K_0, K_1\right)^T$. Inserting the LQR-controller into the equations \eqref{eq:dyn} and \eqref{eq:lindyn} yields the closed-loop dynamics and its linearization:
\begin{align}
\ddot{\theta} &= -\frac{1}{ml^2} \left(mgl\sin(\theta) - K_0\theta - (b+K_1)\dot{\theta}\right) \label{eq:cl} \\
&\simeq -\frac{1}{ml^2} \left((mgl-K_0)\theta - (b+K_1)\dot{\theta}\right). \label{eq:cllin}
\end{align}

Furthermore, consider the actuators torque output to be limited, i.e. $ |u| \leq \bar{u}$. This results in the following equations of motion:
\begin{equation}
\ddot{\theta} = 
\begin{cases}
\frac{1}{ml^2} \left(mgl\sin(\theta) - b\dot\theta - \bar{u} \right) & u(\vec{x})<-\bar{u} \\
\frac{1}{ml^2} \left(mgl\sin(\theta) - b\dot\theta + \bar{u} \right) & u(\vec{x})>\bar{u} \\
\frac{1}{ml^2} \left(mgl\sin(\theta) - K_0\theta - (b+K_1)\dot{\theta} \right) & \text{else}
\end{cases}
\label{eq:full}
\end{equation}

While we know that the origin is globally stable for \eqref{eq:cllin}, for the nonlinear torque limited dynamics (\ref{eq:full}) the stability holds only locally. Thus, we need to estimate the ROA $\mathcal{S}$, in which all states evolve to the origin as time goes to infinity:
\begin{equation}
\mathcal{S} = \left\{ \vec{\gamma} \, \middle\vert \, \vec{x}_{\vec{\gamma}}(t\to\infty)\to 0  \right\}.
\label{eq:act}
\end{equation}

\section{Analytically Derived Region of Attraction Estimation}
\label{sec:ana}
In this section we introduce our approach to estimate the ROA analytically. We start by motivating an inner estimation of $\mathcal{S}$, which depends on the time evolution $u_{\vec{\gamma}}(t)$ with initial value $\vec{\gamma}$. We then approximate $u_{\vec{\gamma}}(t)$ by solving the linear dynamics (\ref{eq:cllin}) and, to compensate for the approximation, add a heuristic constraint. 

\subsection{Effects of Torque Limitations on the Region of Attraction}
An LQR-controller is oblivious to torque limitations. Therefore we expect it to perform well within the limits, but in general it is unclear how the system will behave once the torques are clipped. Especially for safety-critical applications this would be unacceptable. Thus we want to take a conservative route and approximate the ROA as
\begin{equation}
\tilde{\mathcal{S}} = \left\{ \vec{\gamma} \, \middle\vert \, \forall t >0: |u_{\vec{\gamma}}(t)| \leq \bar{u} \right\} \cap \mathcal{S}_\text{unlim},
\label{eq:vio}
\end{equation}
where $u_{\vec{\gamma}}(t)$ is the time evolution of the torque for a system that with initial value $\vec{\gamma}$ and $\mathcal{S}_\text{unlim}$ is the ROA of the system without torque limitations (\ref{eq:cl}). Simulations of (\ref{eq:cl}) show that for the simple pendulum $\mathcal{S}_\text{unlim}$ seems to be the entire phase space. Because of this we can assume $\tilde{\mathcal{S}} \subset \mathcal{S}_\text{unlim}$ and omit $\mathcal{S}_\text{unlim}$ from the following analysis. In the following we will approximate $u_{\vec{\gamma}}(t)$ by explicitly solving the initial value problem for (\ref{eq:cllin}) and inserting the solution into $u(\vec{x})$.

\subsection{Analytic Solution to the Linearized Dynamics}
Solutions to a linear ODE are linear combinations of exponential functions. Their exponents depend on the solution of the characteristic polynomial. In the case of (\ref{eq:cllin}) that is
\begin{equation}
\kappa^2 + \left( \frac{K_1+b}{ml^2} \right) \kappa + \left(\frac{K_0}{ml^2}-\frac{g}{l}\right) = 0,
\end{equation}
with solutions
\begin{equation}
\kappa_{0,1} = \frac{1}{2} \left( -\frac{K_1+b}{ml^2} \pm \sqrt{D}\right),
\end{equation}
where 
\begin{equation}
D = \left(\frac{K_1+b}{ml^2} \right)^2 - 4\left(\frac{K_0}{ml^2}-\frac{g}{l}\right).
\end{equation}

Assuming $D>0$ and $\kappa_{0,1} <0$~\footnote{Due to the LQR-Controller being stable and optimal for the linear dynamics, we expect exponentially decreasing functions and thus negative real-valued exponents.}, the time evolution of $\vec{x}$ is
\begin{equation}
\vec{x}(t) = \left(\begin{array}{rr} 
1 & 1 \\
\kappa_0 & \kappa_1  \\
\end{array}\right) \left(\begin{array}{rr} 
C_0e^{\kappa_0 t} \\
C_1e^{\kappa_1 t} \\
\end{array}\right),
\end{equation}
where
\begin{equation}
\left(\begin{array}{rr} 
C_0 \\
C_1 \\
\end{array}\right) = \frac{1}{\sqrt{D}}\left(\begin{array}{rr} 
-\kappa_1 & 1 \\
\kappa_0 & -1  \\
\end{array}\right) \vec{x}(0).
\end{equation}
 
Inserting $\vec{x}(t)$ into $u(\vec{x})$ yields
\begin{equation}
u_{\text{lin},\vec{x}(0)}(t) = -\left(K_0+K_1\kappa_0\right) C_0e^{\kappa_0 t} -\left(K_0+K_1\kappa_1\right) C_1e^{\kappa_1 t}.
\end{equation}
Let $A_i = -\left(K_0+K_1\kappa_i\right) \kappa_i C_i$. Differentiating $u_{\text{lin},\vec{x}(0)}(t) = 0$ and solving $\dot{u}_{\text{lin},\vec{x}(0)}(t^*) = 0$ for $t^*$, shows that $u(t)$ will have an extremum at
\begin{equation}
t^* = -\frac{ \ln\left(-\frac{A_0}{A_1}\right)}{\sqrt{D}},
\end{equation}
if $-A_0/A_1\geq 0$.

Therefore we can approximate $\tilde{\mathcal{S}}$ as
\begin{align}
\begin{split}
\tilde{\mathcal{S}} \simeq \quad &\left\{ \vec{\gamma} \, \middle\vert \, |u_{\text{lin},\vec\gamma}(0)| \leq \bar{u} \right\} \\
\cap &\left\{ \vec{\gamma} \, \middle\vert \, t^*_{\vec\gamma} > 0 \implies |u_{\text{lin},\vec\gamma}(t^*_{\vec\gamma})| \leq \bar{u} \right\}.
\end{split}
\label{eq:heuroa}
\end{align}

\subsection{Viability of the Linear Approximation}
The linear approximation in the derivations above is only locally viable around the origin, i.e. the pendulums upright position. As a heuristic to do this we consider the difference of the gravity induced torques of the actual and the linearized ODE and compare it to the torque limit. If that difference is smaller than the torque limit, it is reasonable to assume that the controller is still capable to compensate for the error: $mgl | \sin(\theta)-\theta | \leq \bar{u}$. This heuristic only takes the angle as an argument, as the approximation error in this case does not increase with increasing angular velocity.

To conclude, our estimation of the ROA is characterized by three conditions on a given phase space coordinate:
\begin{itemize}
\item It must obey the heuristic $mgl | \sin(\theta)-\theta | \leq \bar{u}$
\item The initial torque must be within the limits $|u_{\text{lin},\vec\gamma}(0)|<\bar{u}$.
\item If the applied torques will exhibit another extremum in the future $t^*>0$, check whether this extremum violates the torque limitations $|u_{\text{lin},\vec\gamma}(t^*)|<\bar{u}$.
\end{itemize}
The resulting region in the phase space is given as
\begin{align}
\begin{split}
\mathcal{S}_{\text{analytic}} =\quad &\left\{ \vec{\gamma} \, \middle\vert \, mgl | \sin(\theta)-\theta | \leq \bar{u} \right\} \\
\cap &\left\{ \vec{\gamma} \, \middle\vert \, |u_{\text{lin},\vec\gamma}(0)| \leq \bar{u} \right\} \\
\cap &\left\{ \vec{\gamma} \, \middle\vert \, t^*_{\vec\gamma} > 0 \implies |u_{\text{lin},\vec\gamma}(t^*_{\vec\gamma})| \leq \bar{u} \right\}.
\end{split}
\label{eq:ana}
\end{align}

\section{Results and Discussion}
\label{sec:exp}
In the following section, we describe and discuss the simulations and experiments in detail. We start by giving an overview of the pendulum's parameters, then we explain the simulations and physical experiments as well as the software implementation\footnote{The open source software implementation can be found at: \url{https://github.com/dfki-ric-underactuated-lab/torque_limited_simple_pendulum/tree/master/software/python/simple_pendulum/controllers/lqr/analytic_roa_estimation}}. Afterwards the results will be presented and interpreted.

\subsection{Pendulum Parameters}
The physical system we used is described in detail in~\cite{article:WB22}, a long exposure shot of it is shown in \FigRef{fig:sys}(b). It has a total mass of $m \approx 0.676$ kg and the distance from the axis of rotation to the center of mass is $l \approx 0.45$ m. Given these parameters the maximum torque that gravity can exert on the system, i.e. at a $90^\circ$ angle, is $\tilde{u} := mlg \approx 2.98$ Nm. The actual motors maximum torque is $6$ Nm. In both physical experiments, we simulated the underactuation by limiting the controllers torque outputs to $\tilde{u}/2 \approx 1.49$ Nm.

For the simulations, we chose three sets of parameters. The `normal' set uses the parameters of the real system. For the other two sets, we varied the length and mass of the pendulum, while keeping $\tilde{u}$ fixed. That way we kept the actuator torques in a reasonable range, but varied the inertial properties of the system. We considered a `long' system with $l/m = 0.1$ and a `short' one with $l/m=10$. The actual parameters and the resulting moments of inertia are given in \TabRef{tab:params}. We also considered three different values of torque limitation $\bar{u} \in \left\lbrace \tilde{u}/2,\ \tilde{u}/4,\ \tilde{u}/8 \right\rbrace$. For both the simulations and physical experiments, the LQRs cost parameters we used are $Q = \text{diag}(1,1)$ and $R = 1$, additionally, we assumed the damping to be around $b \approx 0.1\ \text{Nm s/rad}$. We did not take Coulomb friction into account. We also considered three different values of torque limitation $\bar{u} \in \left\lbrace \tilde{u}/2,\ \tilde{u}/4,\ \tilde{u}/8 \right\rbrace$.

\begin{table}
\caption{Paremeters for Simulation}
\label{tab:params}
\centering
\begin{tabular}{|c|c|c|c|}
\hline
 & mass [kg] & length [m] & inertia [kg m$^2$] \\
\hline
normal & $0.676$ & $0.45$ & $0.137$ \\
\hline
long & $0.174$ & $1.744$ & $0.531$ \\
\hline
short & $1.744$ & $0.174$ & $0.0531$ \\
\hline
\end{tabular}
\end{table}

\subsection{Setup Description}
To evaluate the validity and performance of the analytic ROA estimation $\mathcal{S}_\text{analytic}$ as in (\ref{eq:ana}), we compared it to a Lyapunov-sampling ROA estimation $\mathcal{S}_\text{Lyapunov}$ as a baseline and also put both in the context of the actual ROA $\mathcal{S}$ (see (\ref{eq:act})) and the region of all points that are stabilized without reaching the torque limits $\tilde{\mathcal{S}}$ (see (\ref{eq:vio})).

Since $\mathcal{S}$ and $\tilde{\mathcal{S}}$ cant be computed directly, to illustrate them, we ran $100000$ simulations for each parameter set and with initial values randomly chosen in the range of $-\pi < \theta < \pi$ and $-10\ \text{rad/s} < \omega < 10\ \text{rad/s}$. To integrate (\ref{eq:full}) numerically we used a standard Runge-Kutta algorithm with step width $\Delta t = .1$ s and final time $t_\text{final}=10$ s. We considered the initial values to be in $\mathcal{S}$ if the integration converged to the origin within 5 digit precision. If additionally along the trajectory the torque limits were never exceeded, we considered the initial values to also be in $\tilde{\mathcal{S}}$.

To run similar tests on the physical system, we used a PD-controller to prepare the pendulum at a randomly chosen angle, then applied a torque randomly chosen between $-5$ Nm and $5$ Nm, for $0.5$ s. Afterwards, we switched to the LQR-controller and let the system evolve for $10$ s. We considered the systems state at the time of switching as well as $50$ evenly spaced (timewise) states along the trajectory as initial values $\vec{\gamma}$. Again we recorded if the controller was able to stabilize the pendulum upright ($\mathcal{S}$) and whether or not it met the torque limits during the run ($\tilde{\mathcal{S}}$). In total we conducted $570$ runs resulting $28500$ data points.

In order to analyze the analytical approach, we implemented (\ref{eq:ana}) as an oracle style function $\vec{x} \mapsto \texttt{true}, \text{iff} \ \vec{x} \in \mathcal{S}_\text{analytic}$ by numerically deriving $\mat{K}$ and using it to compute $u_{\text{lin},\vec\gamma}$ and $t^*$. We used this oracle function to estimate for each initial value from simulation and experiment whether or not it lies inside $\mathcal{S}_\text{analytic}$. In addition, to investigate the heuristic used in the previous section, we also implemented an oracle function for $\mathcal{S}_\text{unbound}$ as in (\ref{eq:heuroa}) in an analogous way to the above. 

For $\mathcal{S}_\text{Lyapunov}$, we took $\vec{x}^T\mat{S}\vec{x}$ as a Lyapunov function\footnote{This is the cost-to-go function of the system and known to be a viable Lyapunov function.} and computed the boundary $\rho$ of the largest sublevel set via sampling states and checking the Lyapunov conditions~\cite{article:NB16}. Again we used this to implement an oracle function: $\vec{x} \mapsto \texttt{true}, \text{iff} \ \vec{x}^T\mat{S}\vec{x} \leq \rho$.

To demonstrate the usage of our ROA estimation, we set up a swing-up experiment. We used a combination of an energy-shaping controller to bring the system close to the phase space origin and, once the system is inside the estimated ROA, switch to the LQR to stabilize the pendulum: 
\begin{equation}
u(\vec{x}) = \begin{cases}
-\mat{K}\vec{x}, & \text{if} \ \vec{x} \in \mathcal{S}_\text{analytic} \\
-c\dot{\theta}\Delta E(\vec{x}) + b\dot{\theta}, & \text{else}
\end{cases},
\label{eq:swi}
\end{equation}
with a positive constant $c$ and the difference in total energy w.r.t. origin $\Delta E(\vec{x})=mgl(\cos(\theta)-1)-\frac{1}{2}ml^2\dot{\theta}^2$.

\subsection{Validation of Heuristic}
\begin{figure}
\centering
\includegraphics[width=.99\columnwidth]{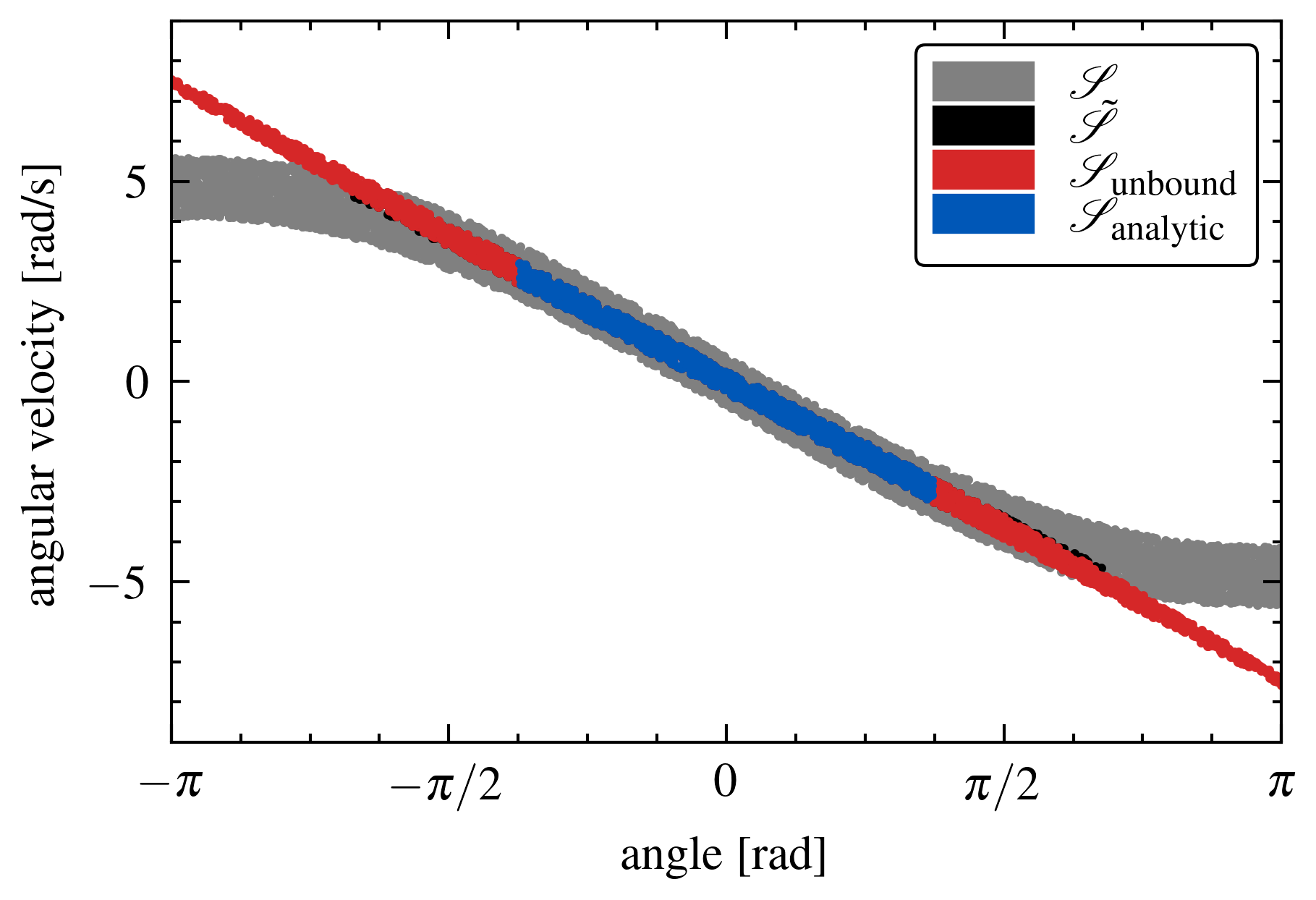}
\caption{Effect of the heuristic on the ROA. Depicted are the analytic ROA estimation both with ($\mathcal{S}_\text{analytic}$) and without heuristic bounds ($\mathcal{S}_\text{unbound}$). As reference also $\mathcal{S}$ and $\tilde{\mathcal{S}}$ from the simulation are shown.}
\label{fig:heu}
\end{figure}

To evaluate the validity of the heuristic added in (\ref{eq:ana}), we compared the simulation results to the analytic ROA estimation with and without taking the heuristic into account. \FigRef{fig:heu} illustrates the results. Here we plotted all four regions for the `long' configuration with $\bar{u} = \tilde{u}/2$ as an example. The grey and black regions depict $\mathcal{S}$ and $\tilde{\mathcal{S}}$ respectively, the blue region is the analytic result $\mathcal{S}_\text{analytic}$ and in red the region $\mathcal{S}_\text{unbound}$ without the heuristic bound is drawn.

Comparing $\mathcal{S}_\text{unbound}$ with $\mathcal{S}$ in the upper left and lower right parts of the plot, we find that a significant number of initial values are within $\mathcal{S}_\text{unbound}$ but have failed to stabilize in simulation that means $\mathcal{S}_\text{unbound}$ is not a conservative estimate of the ROA. Therefore, a heuristic to bound our estimation is indeed necessary.

To ensure the applied heuristic is sufficient, we searched in all initial values evaluated by simulation for false positives of the kind $\vec{\gamma} \in \mathcal{S}_\text{analytic} \setminus \mathcal{S}$ and found none. Thus, it seems to be a sufficient choice, but \FigRef{fig:heu} also shows that the heuristic bound reduces the area of the estimation by a lot.

\subsection{Simulation Results}
\label{subs:num}
\begin{figure*}
\includegraphics[width=\textwidth]{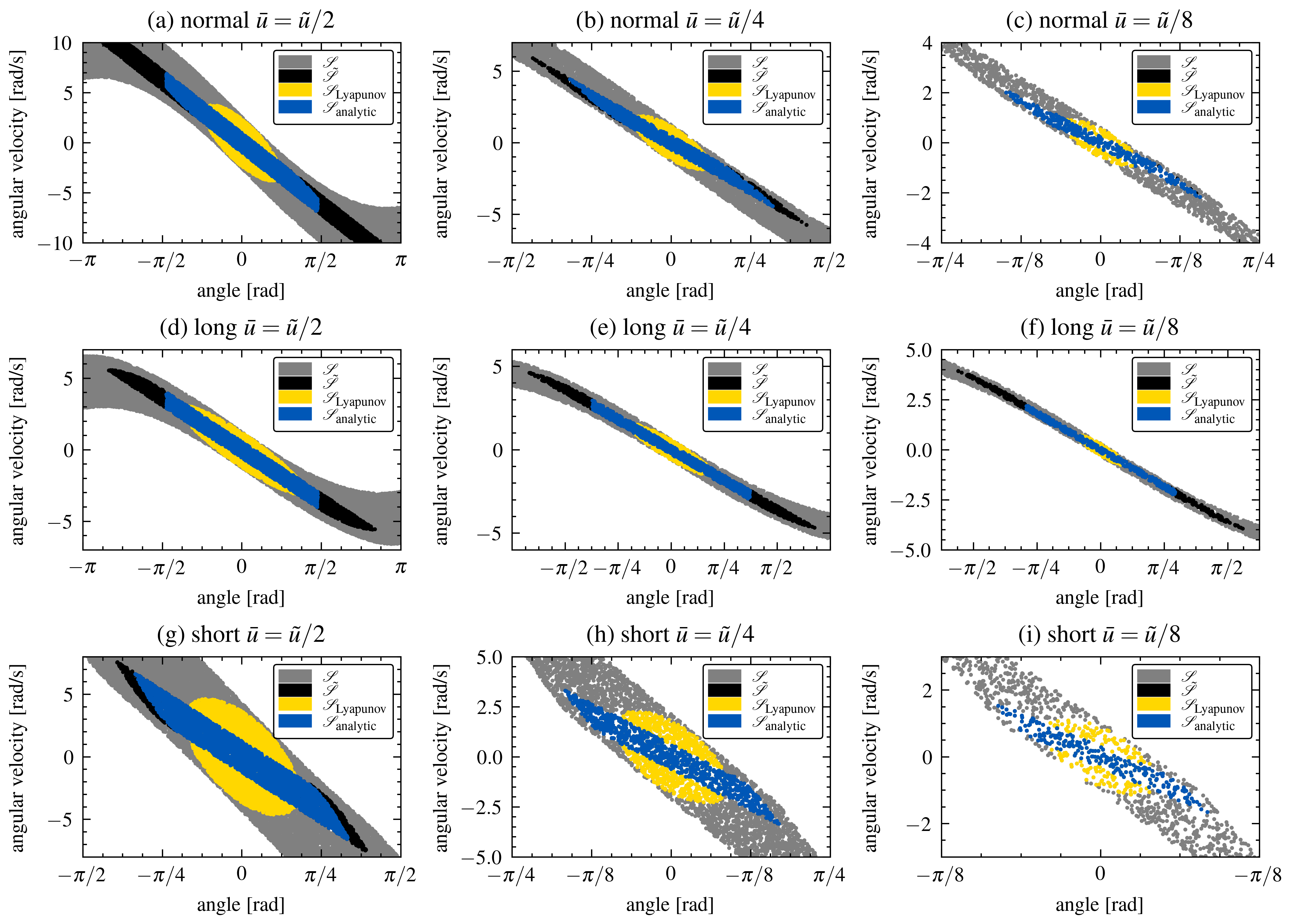}

\caption{Comparison of analytic ROA ($\mathcal{S}_\text{analytic}$) to the Lyapunov baseline ($\mathcal{S}_\text{Lyapunov}$). As reference also $\mathcal{S}$ and $\tilde{\mathcal{S}}$ from the simulation are shown. The systems mass and length vary between subfigures according to \TabRef{tab:params} and the torque limit $\bar{u}$ varies as noted in the subcaptions.}
\label{fig:num}
\end{figure*}

As an overview of the results from simulation, in \FigRef{fig:num}, we plotted $\mathcal{S}_\text{analytic}$ as blue, $\mathcal{S}_\text{Lyapunov}$ as yellow and, again, $\mathcal{S}$ and $\tilde{\mathcal{S}}$ as black and grey areas respectively. The subplots show the results for each set of parameters and each torque limit considered. From looking at \FigRef{fig:num}, we find that neither the analytically estimated ROA is a superset of the opimization based ROA nor the other way around. So none of the approaches seem to be clearly superior regarding the ROA's area. Due to their distinct difference in shape, they do not overlap too much, therefore it might in fact be practical to combine both methods for a larger overall ROA.

For a quantitative comparison of the areas, we estimated the relative area of $\mathcal{S}_\text{analytic}$ from the numbers of initial values inside each ROA as $A_{rel} \simeq N_{analytic}/N_{Lyapunov}$ for each parameter set and torque limit. The results are given in \TabRef{tab:relvol}. They show that the analytic ROA's area is bigger than the Lyapunov based one in four cases, namely the `normal' and the `long', where $\bar{u} \leq \tilde{u}/4$. For the `short' system the Lyapunov-sampling estimation is strictly better. The numbers suggest that the analytic approach gets better as the torque limits get more severe. In our experiments we found only one exception to that rule. From `normal' $\tilde{u}/4$ to $\tilde{u}/8$ the relative area decreased, but this might be explained with statistical errors increasing with the smaller sample sizes in smaller regions.

\begin{table}
\caption{Relative Volume of the Analytic ROA Estimation}
\label{tab:relvol}
\centering
\begin{tabular}{|c|c|c|c|}
\hline
 & $\tilde{u}/2$ & $\tilde{u}/4$ & $\tilde{u}/8$ \\
\hline
normal & $0.99$ & $1.183$ & $1.150$ \\
\hline
long & $0.723$ & $1.152$ & $1.721$ \\
\hline
short & $0.718$ & $0.72$ & $0.767$ \\
\hline
\end{tabular}
\end{table}

As an additional aspect, we measured computation times for the oracle functions of both $\mathcal{S}_\text{analytic}$ and $\mathcal{S}_\text{Lyapunov}$, i.e. the time it took to prepare the function for a new set of parameters, not the evaluation for a given initial value. The mean of the results are:
\begin{align*}
&T_\text{analytic} = 0.00136 \text{ s} 
&T_\text{Lyapunov} = 12.8 \text{ s}
\end{align*}
To put these numbers in perspective: computing $\mathcal{S}$ and $\tilde{\mathcal{S}}$ took about $5$ hours for a single set of parameters. And to check the stability of states that were not numerically computed, the use of convex hull approximation of the point cloud would be required.

It took about $10^5$ times as long to compute the Lyapunov based prediction as it did for the analytic method. This could be very relevant for robotic applications where inertial parameters vary, such as grasping or carrying external objects. For $\mathcal{S}_\text{Lyapunov}$ we sampled $500$ times. Thus, even reducing this number to where the sampling stops converging properly, a difference of $3$ to $4$ orders of magnitude between $T_\text{analytic}$ and $T_\text{Lyapunov}$ would remain. However, one has to keep in mind that the analytic method only yields an oracle style prediction function, whereas the Lyapunov method yields a conservative geometrically meaningful description of the estimated ROA in addition, which is useful in many cases.

\subsection{Experimental Results}
\begin{figure}
\centering
\includegraphics[width=.99\columnwidth]{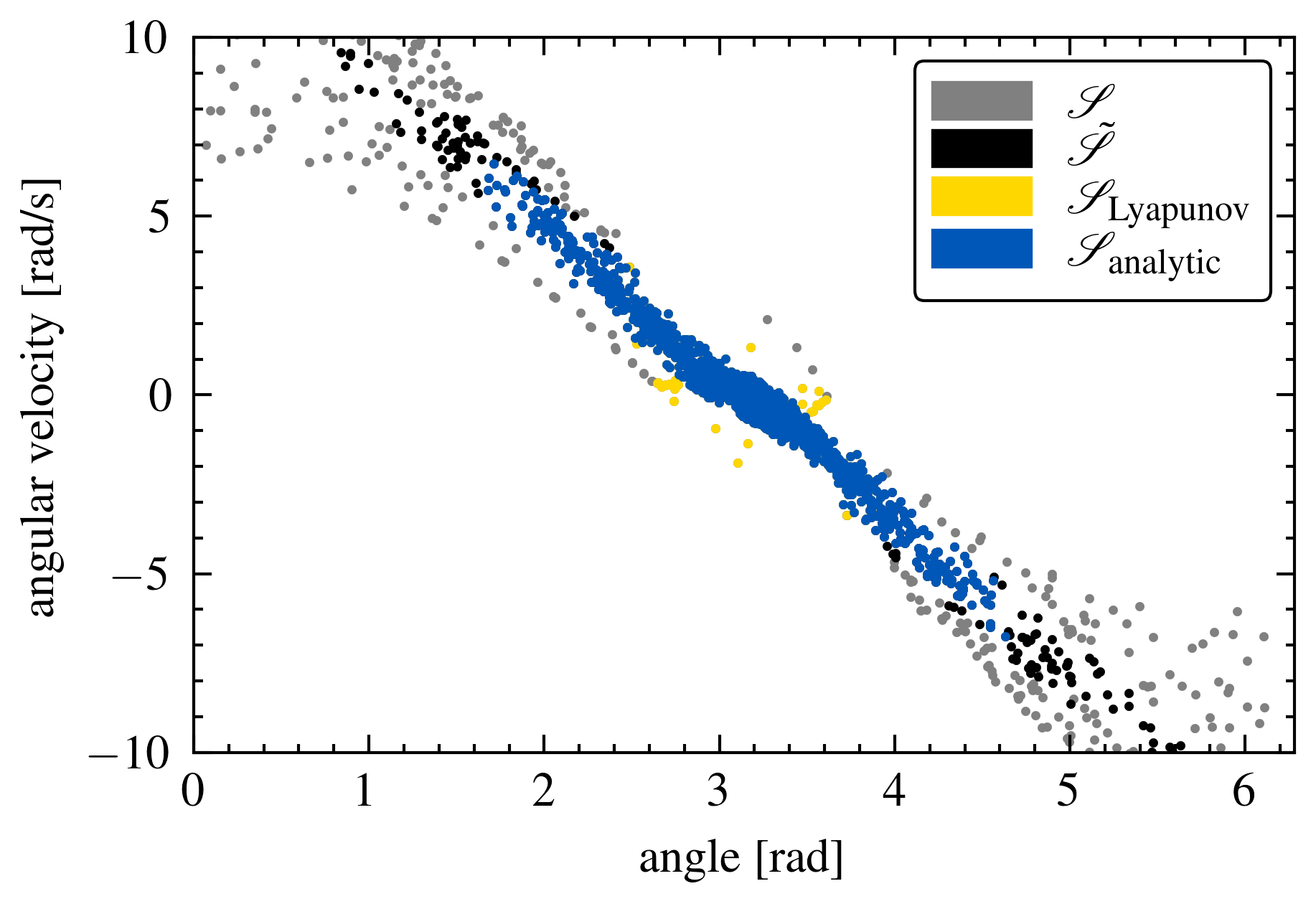}
\caption{Experimental evaluation of estimated ROA. Depicted are the analytic ROA estimation ($\mathcal{S}_\text{analytic}$), the Lyapunov based method ($\mathcal{S}_\text{Lyapunov}$), and the experimentally derived $\mathcal{S}$ and $\tilde{\mathcal{S}}$. The `normal' parameters with $\bar{u} = \tilde{u}/2$ were used.}
\label{fig:exp}
\end{figure}

The experimental results draw a similar picture to the simulations. As can be seen in \FigRef{fig:exp}, the experimentally derived regions look very similar to the ones from \FigRef{fig:num} (a). Most notably, all states inside an estimated ROA, both analytic and Lyapunov-sampling, were stabilized by the controller. This suggests that the systems dynamics were modeled well, and thus, it is reasonable to use our ROA estimation on a real system.

\label{subs:phy}

\begin{figure}
\centering
\includegraphics[width=.99\columnwidth]{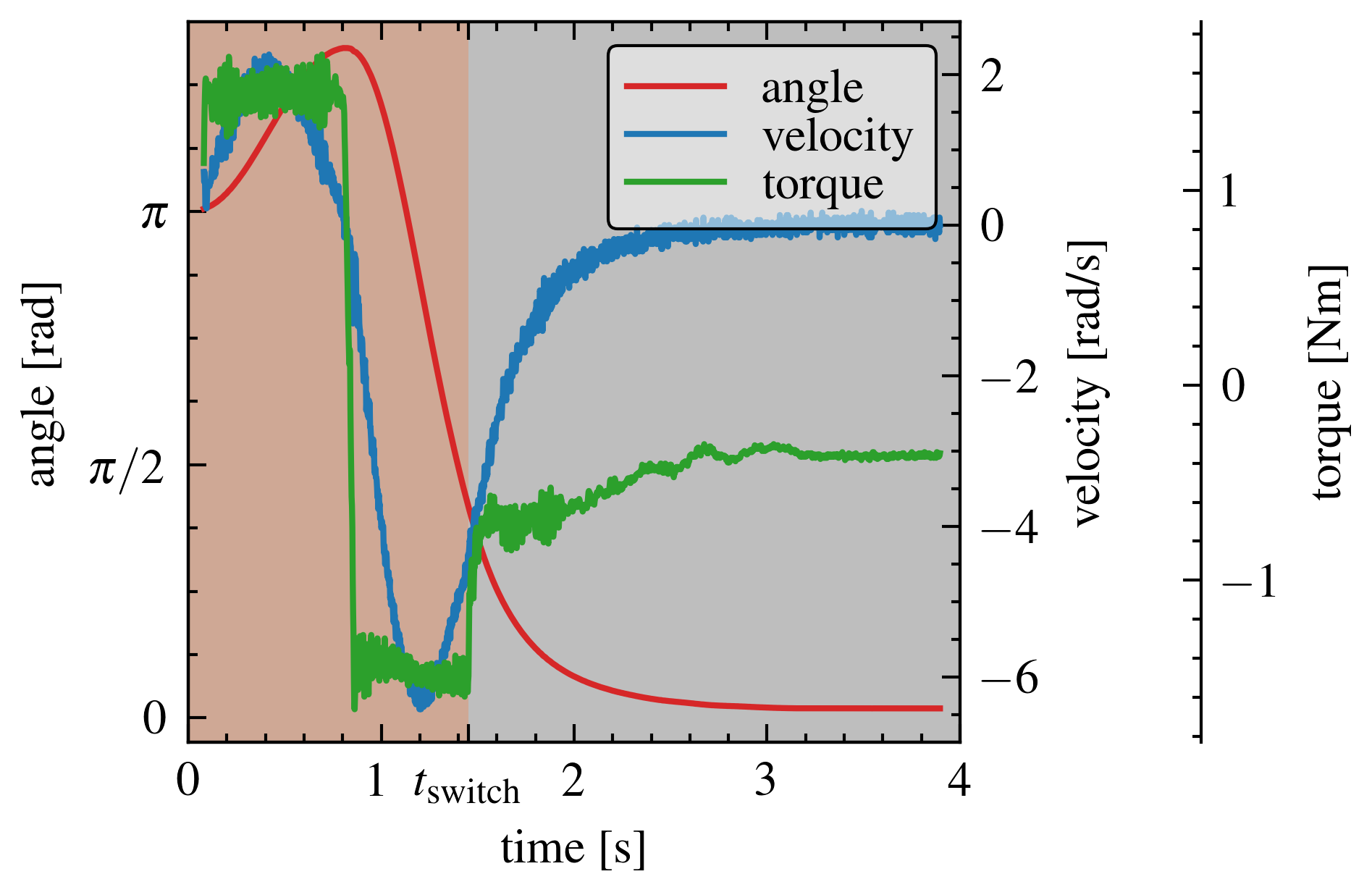}
\caption{Swing-up of pendulum with imposed torque limit of $\bar{u} = \tilde{u}/2$. A combination of an energy-shaping for swing-up and an LQR-controller for stabilizing the upright position was used. The swing-up controller is active in the brown and the LQR in the gray area. The switching behavior was implemented using the analytically estimated ROA.}
\label{fig:swi}
\end{figure}

In \FigRef{fig:swi} the angle, velocity, and torque during the swing-up experiment are shown. We see that the controller (\ref{eq:swi}) managed to move the pendulum from the lower fixed point to the upright position. In fact close inspection shows that the angle did not quite reach the origin. This, as well as the non vanishing torques is probably due to friction canceling out the controller torque. 

The system reached the estimated ROA at $t_\textbf{switch} \approx 1.45$ s, $\vec{x} \approx (0.416\pi\text{ rad}, -4.4\text{ rad/s})$. A look at \FigRef{fig:num} (a) shows that the controller would not have switched at that time, if the Lyapunov based approach would have been used instead. Thus, at least for this single run, our ROA estimation provided a better prediction.

\section{Conclusion and Outlook}
\label{sec:con}
To summarize, we introduced an analytic approach to ROA estimation for an LQR-controlled simple pendulum with torque limitations. We implemented this approach and compared it to a baseline Lyapunov-sampling method in both simulation and physical  experiments. In the first case we varied the inertial parameters of the pendulum as well as the enforced torque limitation. In addition, we successfully demonstrated a swing-up controller that switches between energy-shaping and LQR based on our approach as an exemplary use-case.

As far as our simulation results go, the new analytically estimated ROA is between $0.7$ and $1.7$ times the size of the Lyapunov baseline. So generally the performance is similar. Our results suggest that the analytic approach becomes better with more severe torque limitations. Maybe more importantly the two estimations overlap is comparatively small, so it might be of interest to experiment with combining both methods. In addition to that, given a new set of parameters, it was about $10^5$ times faster to compute an oracle prediction function analytically than with the Lyapunov approach. The experimental ROAs showed similar results. Most notably, even on the real system, all states estimated to be inside the ROA were in fact stabilized by the LQR-controller. 

One rather obvious limitation of our approach is that all calculations done in Section~\ref{sec:ana} are to some degree specific to the pendulums dynamics and higher dimensional systems might be more intricate to solve. However, linear ODEs are quite well studied and solutions to higher orders are known as well~\cite{book:BR91},~\cite{book:Robinson04}, so additional research might lead to a more general version of this analytic method. Open problems regarding more complex systems include finding a heuristic where the linear approximation breaks, as well as estimating the ROA of the system without torque limitations $\mathcal{S}_\text{unlim}$.

Another possible extension would be to find solutions to the linearization of the composed ODE from (\ref{eq:full}). In principle one could solve the components individually and then try to find the states and times of the transitions from the initial value problem. This way one could include initial values into the ROA estimation, which start out in the area of saturated torque, as long as they end up in the controllable area.

As mentioned in the introduction, one motivation for this work was to enable the application of formal verification methods to the problem of ROA estimation. A possible first step towards that goal would be to translate this analytic approach, or parts of it, into a rigorous logic framework. A promising candidate here would be differential dynamic logic~\cite{article:Platzer17} and the associated prover keymaeraX~\cite{inproceedings:MQVP15}. Here, the use of a heuristic to bound the linear approximations viability might become problematic. Further investigations towards a better understanding of this is probably necessary.







\bibliographystyle{IEEEtran}
\bibliography{IEEEexample,bibliography}

\end{document}